# Machine Recognition of Hand Written Characters using Neural Networks

Yusuf Perwej
Department of Computer Science
Singhania University, Rajsthan, India

Ashish Chaturvedi
Department of Applied Sciences
Gyan Bharti Institute of Technology, Meerut, India

## ABSTRACT
Even today in Twenty First Century Handwritten communication has its own stand and most of the times, in daily life it is globally using as means of communication and recording the information like to be shared with others. Challenges in handwritten characters recognition wholly lie in the variation and distortion of handwritten characters, since different people may use different style of handwriting, and direction to draw the same shape of the characters of their known script. This paper demonstrates the nature of handwritten characters, conversion of handwritten data into electronic data, and the neural network approach to make machine capable of recognizing hand written characters.

## 1. INTRODUCTION
Handwritten characters are vague in nature as there may not always be sharp perfectly straight lines, and curves not necessarily be smooth, unlikely the printed characters. Furthermore, characters can be drawn in different sizes and orientation which are often supposed to be written on a baseline in an upright or downright position. Therefore, a robust handwritten recognition system can be designed by considering these limitations. However, it is quiet tedious sometimes to recognize hand written characters as it can be seen that most of the people can not even read their own written notes. Therefore, there is an obligation for a writer to write clearly. But even today in Twenty First Century Handwritten communication has its own stand and most of the times, in daily life it is globally using as means of communication and recording the information like to be shared with others.

Researchers already paid many efforts in designing hand written character recognition system most of them cited as [1-5] because of its important application like bank checking process, reading postal codes and reading different forms [6]. Handwritten digit recognition is still a problem for many languages like Arabic, Farsi, Chinese, English, etc [7]. A machine can perform more tasks than a human being in the same time; this kind of application saves time and money and eliminates the requirement that a human perform such a repetitive task. For the recognition of English handwritten characters, various methods have been proposed [8-12]. Also a few numbers of studies have been reported for Farsi language [13-15].

Proposed Hand written character recognition system for machine recognition can be developed in these phases: scanning of hand written characters i.e conversion into electronic data, usually an black & white image file; some preprocessing can be applied to the image; then the feature of the character will be extracted from the image; finally, on the basis of extracted features from the image, the character can be classify to recognize using gradient descent learning method for feed forward neural network. In next sections we explore the proposed hand written character recognition system step by step. Finally, in the last section results will be discussed which are obtained from the system and conclusion will be made on the basis of result obtained.

## 2. PROPOSED HANDWRITTEN CHARACTER RECOGNITION SYSTEM
The construction of Handwritten Character Recognition system consists of several phases. First phase is to preprocessing the string of characters. Then second phase is to extract the features from the string of characters. It is a very important phase. This feature extraction method must very effective and efficient since the extracted features will become the basis of recognition. Thus the feature extraction will be done at the unit level i.e. for individual characters bounding boxes in the string and finally the union operator can provide the sum of features for the string of characters. These extracted features will then provided as input patterns to neural networks system. Once the neural network system has been trained for these input patterns, it will be able to classify them.

### 2.1 Preprocessing phase
In this phase each character in string is pre-processed. It deals with technique for enhancing contrast; removing noise and isolating regions whose texture indicate a likelihood of character information. In preprocessing phase it is being normalized and removing all redundancy errors from the character image and sends it to next stage of feature extraction.
The following preprocessing steps are followed here to make the individual characters smooth and clear:
i) Firstly, the characters are cropped i.e. extra pixels are removed from the character image.
ii) Then, RGB characters image is converted into Gray scale image.
iii) After that, edges are finding out of the individual characters.
iv) Extra holes fill up from the characters.
v) Bounding Boxes are made up of all the individual characters in the string of hand written characters. These boxes represent the area of individual character. We have tested our experiment over hand written Hindi characters and their strings.

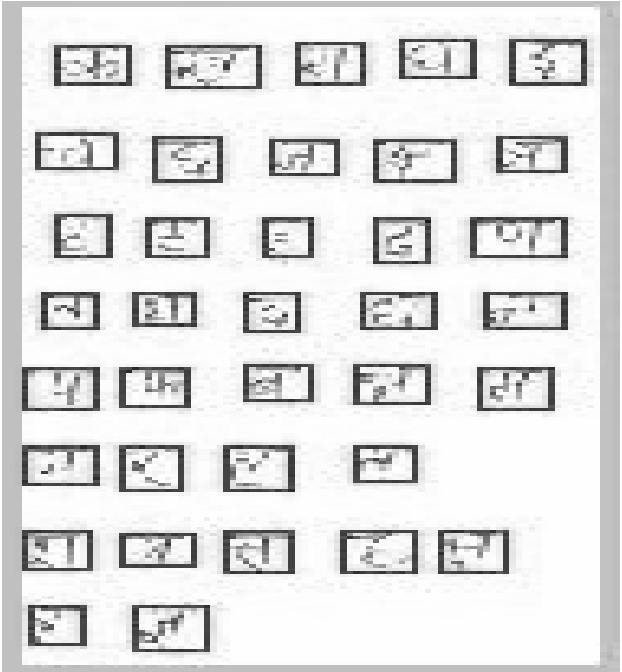

**Figure 1 representing Bounding area of a character**

## 2.2 Feature Extraction Phase

Transforming the input data into the set of features is called features extraction. Feature extraction is a special characteristic of dimensionality reduction methods. Analysis with a large number of variables generally requires a large amount of memory and computation power or a classification algorithm, which over fits the training sample and generalizes poorly to new samples. When the input data is too large to be processed then the input data will be transformed into a reduced representation set of features. Obviously, it is important to select type of feature extraction method because it is the important, factor in the performance of pattern recognition systems [16]. Selection type of feature extraction is dependent on the application. Different features are purposed to recognize hand written digits and characters. They are Furies Transform, Invariant Moments, Geometric Moments, Characteristic Loci and others [17]-[22]. In this paper we are using Characteristic Loci to recognize handwritten characters.

It is already been established that the Characteristic Loci Feature extraction method gives good results in handwritten digit recognition [23]. Characteristic Loci Feature extraction method is commonly in vertical, horizontal and 45 or 135 degree orientations. A digit is related to each point of picture. This digit is dependent on the number of contacts along four directions, up, down, left and right. This feature vector has 81 components that each of them is relative amplitude of the specified digit in the picture. To normalize this, it can be divided to the number of black pixels.

One of the most important transform to reduce dimensionality for simple and fast data processing and picture classification is Principal component analysis (PCA). Now it is mostly used as a tool in exploratory data analysis and for making predictive models. It involves a mathematical procedure that transforms a number of possibly correlated variables into a smaller number of uncorrelated variables called principal components. There are three reasons why PCA is an appropriate transformation method for the handwritten characters. Firstly, PCA as a linear transformation method is very simple. Secondly, the components of the transformed feature vector are statistically independent. Another reason for using PCA is that the feature components are ordered according to their importance. By removing the last some n components, the dimensionality of the feature vector may be reduced without losing too much information. The optimal value for n can be empirically determined for different applications. Suppose (feature vector) of train data are $P_1, P_2, P_3 \ldots P_N$.

$$A = \frac{1}{N} \sum_{n=1}^{N} P_n \quad (1)$$

$$X_j = P_j - A \quad 1 < j < N \quad (2)$$

$$Y = X_1 X_2 \ldots \ldots X_N \quad (3)$$

Where A is the mean vector, Xi is the difference between feature vector and the mean vector and Y matrix is determined by Equation (3). Covariance Matrix is M×M matrix that is represented in Equation (4); M is the dimension of feature vector

$$A = \frac{1}{N} \sum_{n=1}^{N} X_n X_n^T = \frac{1}{N} YY^T \quad (4)$$

In this paper, first pictures are mapped in to center of a zero matrix. Then all of pictures are the same size. The component of characteristic Loci features was calculated. Then a few number of PCA component was selected in those. These feature vector are transform matrices. These reduced vectors are given to neural network. By this method feature vector is reduced and large computational is decreased.

## 2.3 Back propagation based Neural network recognition system

The back-propagation algorithm is a gradient descent method minimizing the mean square error between the actual and target output of a multilayer perceptron. Assuming sigmoidal nonlinear function

$$f(net_i) = \frac{1}{1 - e^{-net}} \quad (5)$$

The back-propagation algorithm consists of the following steps:

**I. Initialize Weights and Offsets**
Initialize all weights and node offsets to small random values.
**II. Present Input and Desired Output Vector**
Present continuous input vector **x** and specify the desired output **d**. Output vector elements are set to zero values except for that corresponding to the class of the current input.
**III. Calculate Actual Outputs**
Calculate the actual output vector **y** using the sigmoidal nonlinearity.
**IV. Adapt weights**
Adjust weights by

$$w_{ij}(t+1) = w_{ij}(t) + \iota \delta_j x_i^{'} \quad (6)$$

where is the output of the node i and is the sensitivity of the node j. If node j is an output node, then

$$\delta_j = f'(net_j)(d_j - y_j) \quad (7)$$

where dj is the desired output of the node j, yj is the actual output and is the derivation of the activation function calculated at netj. If the node j is an internal node, then the sensitivity is defined as

$$\delta_j = f'(net_j) \sum_k \delta_k w_{jk} \qquad (8)$$

where k sums over all nodes in the layer above the node j. Update equations are derived using the chain derivation rule applied to the LMS training criterion function. Convergence can be faster if a momentum term is added and weight changes are smoothed by

$$w_{ij}(t+1) = w_{ij}(t) + \iota \delta_j x_i' + \alpha [w_{ij}(t) - w_{ij}(t-1)] \qquad (9)$$

a) The input pattern file
b) No. of neurons in each hidden layer
c) Value of learning rate
d) Value of momentum constant
e) Error value for convergence

The output of training program is a file which contains modified weights of different connection of the network. This file is used as the input to testing program. This file also contains the values of numbers of neurons in input layer, Hidden layers, output layer, value of learning rate and momentum factor so that user is no further required to re-enter these values during testing.

After the completion of the training phase, a test pattern is given to the neural network and the results are compared with the desired result. Difference between the two values gives the error. Percentage accuracy can be obtained as:

% Accuracy = <u>No of characters found correctly</u> *100
                 Total no of patterns

| Characters string | No. of samples for training | No. of samples for testing | No. of hidden units | No. of epochs | % Recognition Accuracy |
|---|---|---|---|---|---|
| (त श ड) | 100 | 25 | 30 | 700 | 92.0 |
| (श ख सा) | 140 | 25 | 36 | 888 | 95.0 |
| (ड ट ण) | 120 | 25 | 36 | 920 | 91.2 |
| (ह न प) | 90 | 25 | 36 | 799 | 98.5 |
| (श ख सा त श ड) | 140 | 25 | 48 | 1010 | 88.0 |
| (ड ट ण ह न प) | 140 | 25 | 54 | 1200 | 94.5 |
| (त श ड ह न प) | 120 | 25 | 60 | 1112 | 96.0 |
| (श ख सा ड ट ण) | 100 | 25 | 54 | 1220 | 90.1 |
| (ड ट ण त श ड) | 130 | 25 | 60 | 1030 | 96.5 |
| (श ख सा ह न प) | 95 | 25 | 60 | 990 | 93.4 |

**V. Repeat by Going to Step 2**
The program based on the back propagation algorithm as described above, trains the network to recognize the handwritten characters. This network takes input-output vector pairs during training. The network trains its weight array to minimize the selected performance measure, i.e., error using back propagation algorithm.

Our designed neural network system takes following as inputs from the user:

## 3. RESULTS & DISCUSSION

First of all, training of system is done by using different data set or samples and then system is tested for few of the given samples, and accuracy is measured. The data set was partitioned into two parts. The first part is used for training the system and the second was for testing purpose. For each character, feature were computed and stored for training the network. Following parameters are used for training of our Neural Network system:

No. of neurons in Input Layer: 6
No of neurons in Hidden Layer: 8
Transfer Function Used for Layer 1: "Logsig"
Transfer Function Used for Layer 2: "Tansig"
Adaption Learning Function: "Learngdm"
Performance Function: "MSE"

The table given below display the results obtained from experiment.

As we can observe from the table that as we increase the number of units in hidden layer, network converges with more accuracy and this can be seen for the whole pattern. The recognition accuracy rate is very up to the mark and as per our predictions. The experiment shows that the string of handwritten characters can be recognized by the machine with a significant accuracy.

## 4. CONCLUSION

We have proposed and developed a scheme for recognizing hand written characters. We have tested our experiment over Hand written Hindi characters and the strings of these Hand written characters. Experimental results shown that the machine has successfully recognized the characters string with the average accuracy of 93.5%, which significant and may be acceptable in some applications. The experiment shows as we have increased the number of hidden units in hidden layers, the network converges with more accuracy. In future the similar experiment can be tested over some other characters and with some more or new parameters to improve the accuracy of the machine.